\documentclass{article}

\usepackage[preprint]{nips_2018}

\usepackage[utf8]{inputenc} % allow utf-8 input
\usepackage[T1]{fontenc}    % use 8-bit T1 fonts
\usepackage{hyperref}       % hyperlinks
\usepackage{url}            % simple URL typesetting
\usepackage{booktabs}       % professional-quality tables
\usepackage{amsfonts}       % blackboard math symbols
\usepackage{nicefrac}       % compact symbols for 1/2, etc.
\usepackage{microtype}      % microtypography
\usepackage{comment}

\usepackage{longtable}
\usepackage{todonotes}
\usepackage{caption}
\usepackage{subcaption}

\usepackage{amsthm}
\usepackage{amsmath}
\usepackage{amsfonts}
\usepackage{amssymb}
\usepackage{comment}
\usepackage{csquotes}
\usepackage[framemethod=tikz]{mdframed}

\usepackage{appendix}
\usepackage{chngcntr}
\usepackage{etoolbox}

\usepackage[ruled]{algorithm2e}
\usepackage{algpseudocode}

\usepackage{lettrine}

\usepackage{mathtools}% http://ctan.org/pkg/mathtools
\usepackage{amsmath}

\usepackage{tabularx}

\usepackage{listings}
\usepackage{color}
\definecolor{mygreen}{rgb}{0,0.6,0}
\definecolor{mygray}{rgb}{0.5,0.5,0.5}
\definecolor{mymauve}{rgb}{0.58,0,0.82}

\newcommand{\bb}[1]{\mathbf{#1}}
\newcommand{\bbb}{\bb{b}}

\newcommand{\bx}{\bb{x}}

\newcommand{\bt}{\bb{t}}
\newcommand{\by}{\bb{y}}

\newcommand{\bW}{\bb{W}}

\newcommand{\bg}{\bb{g}}
\newcommand{\bh}{\bb{h}}

\newcommand{\bz}{\bb{z}}

\newcommand{\bbf}{\bb{f}}

\newcommand{\bT}{\boldsymbol{\theta}}

\newcommand{\bs}{\mathbf{s}}

\newcommand{\pT}{p_{\bT}}

\newcommand{\fT}{\bbf_{\bT}}
\newcommand{\gT}{\bg_{\bT}}

\title{
Glow: Generative Flow\\with Invertible 1$\times$1 Convolutions
}

\author{Diederik P. Kingma\textsuperscript{*}, Prafulla Dhariwal\thanks{Equal contribution.}\\
OpenAI, San Francisco}

\begin{document}

\maketitle

\begin{abstract}
Flow-based generative models ~\citep{dinh2014nice} are conceptually attractive due to tractability of the exact log-likelihood, tractability of exact latent-variable inference, and parallelizability of both training and synthesis. In this paper we propose \emph{Glow}, a simple type of generative flow using an invertible $1 \times 1$ convolution. Using our method we demonstrate a significant improvement in log-likelihood on standard benchmarks. Perhaps most strikingly, we demonstrate that a generative model optimized towards the plain log-likelihood objective is capable of efficient realistic-looking synthesis and manipulation of large images. The code for our model is available at \url{https://github.com/openai/glow}.
\end{abstract}

\section{Introduction}

\begin{comment}
 - Put paper into context. Explain importance of generative modeling.
 - Explain likelihood-based generative models.
 - Explain (in words) what sets NICE/RealNVP type models apart from other likelihood-based models.
 - Summarize our contribution and its importance.
 - Add cool-looking figure of generated faces to lure the reader.
\end{comment}

\begin{figure}[!b]
	\centering
	\includegraphics[width=0.8\textwidth]{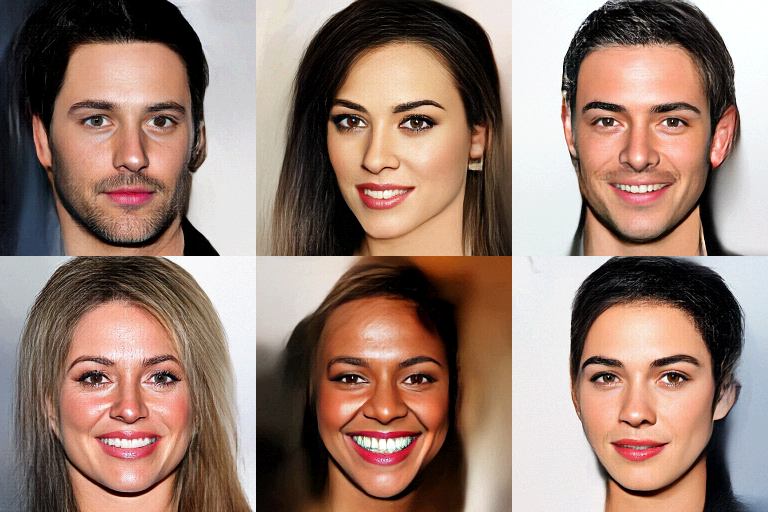}
	\caption{Synthetic celebrities sampled from our model; see Section~\ref{sec:method} for architecture and method, and Section~\ref{sec:experiments} for more results.}
	\label{fig:xor}
\end{figure}

Two major unsolved problems in the field of machine learning are (1) data-efficiency: the ability to learn from few datapoints, like humans; and (2) generalization: robustness to changes of the task or its context. AI systems, for example, often do not work at all when given inputs that are different from their training distribution. A promise of \emph{generative models}, a major branch of machine learning, is to overcome these limitations by:
(1) learning realistic world models, potentially allowing agents to plan in a world model before actual interaction with the world, and (2) learning meaningful features of the input while requiring little or no human supervision or labeling. Since such features can be learned from large unlabeled datasets and are not necessarily task-specific, downstream solutions based on those features could potentially be more robust and more data efficient. In this paper we work towards this ultimate vision, in addition to intermediate applications, by aiming to improve upon the state-of-the-art of generative models.

Generative modeling is generally concerned with the extremely challenging task of modeling all dependencies within very high-dimensional input data, usually specified in the form of a full joint probability distribution. Since such joint models potentially capture all patterns that are present in the data, the applications of accurate generative models are near endless. Immediate applications are as diverse as speech synthesis, text analysis, semi-supervised learning and model-based control; see Section~\ref{sec:relatedwork} for references.

The discipline of generative modeling has experienced enormous leaps in capabilities in recent years, mostly with likelihood-based methods~\citep{graves2013generating,kingma2013auto,kingma2018vaes,dinh2014nice,pixelrnn} and generative adversarial networks (GANs)~\citep{goodfellow2014generative} (see Section ~\ref{sec:relatedwork}). Likelihood-based methods can be divided into three categories:
\begin{enumerate}
\item Autoregressive models~\citep{hochreiter1997long,graves2013generating,pixelrnn,oord2016conditional,van2016wavenet}. Those have the advantage of simplicity, but have as disadvantage that synthesis has limited parallelizability, since the computational length of synthesis is proportional to the dimensionality of the data; this is especially troublesome for large images or video.
\item Variational autoencoders (VAEs) ~\citep{kingma2013auto,kingma2018vaes}, which optimize a lower bound on the log-likelihood of the data. Variational autoencoders have the advantage of parallelizability of training and synthesis, but can be comparatively challenging to optimize~\citep{kingma2016improving}.
\item Flow-based generative models, first described in NICE~\citep{dinh2014nice} and extended in RealNVP~\citep{dinh2016density}. We explain the key ideas behind this class of model in the following sections.
\end{enumerate}

Flow-based generative models have so far gained little attention in the research community compared to GANs~\citep{goodfellow2014generative} and VAEs~\citep{kingma2013auto}. Some of the merits of flow-based generative models include:
\begin{itemize}
\item Exact latent-variable inference and log-likelihood evaluation. In VAEs, one is able to infer only approximately the value of the latent variables that correspond to a datapoint. GAN’s have no encoder at all to infer the latents. In reversible generative models, this can be done exactly without approximation. Not only does this lead to accurate inference, it also enables optimization of the exact log-likelihood of the data, instead of a lower bound of it.
\item Efficient inference and efficient synthesis. Autoregressive models, such as the PixelCNN~\citep{oord2016conditional}, are also reversible, however synthesis from such models is difficult to parallelize, and typically inefficient on parallel hardware. Flow-based generative models like Glow (and RealNVP) are efficient to parallelize for both inference and synthesis.
\item Useful latent space for downstream tasks. The hidden layers of autoregressive models have unknown marginal distributions, making it much more difficult to perform valid manipulation of data. In GANs, datapoints can usually not be directly represented in a latent space, as they have no encoder and might not have full support over the data distribution. ~\citep{grover2018flow}. This is not the case for reversible generative models and VAEs, which allow for various applications such as interpolations between datapoints and meaningful modifications of existing datapoints.
\item Significant potential for memory savings. Computing gradients in reversible neural networks requires an amount of memory that is constant instead of linear in their depth, as explained in the RevNet paper~\citep{gomez2017reversible}.
\end{itemize}

In this paper we propose a new a generative flow coined \emph{Glow}, with various new elements as described in Section~\ref{sec:method}. In Section~\ref{sec:experiments}, we compare our model quantitatively with previous flows, and in Section ~\ref{sec:qualexperiments}, we study the qualitative aspects of our model on high-resolution datasets.

\section{Background: Flow-based Generative Models}\label{sec:flow}
\begin{comment}
 - Explain mathematics behind NICE / RealNVP
 - Explain benefits
 - Explain downsides: potential downsides of bipartite structure
\end{comment}

Let $\bx$ be a high-dimensional random vector with unknown true distribution $\bx \sim p^*(\bx)$.
We collect an i.i.d. dataset $\mathcal{D}$, and choose a model $\pT(\bx)$ with parameters $\bT$. In case of discrete data $\bx$, the log-likelihood objective is then equivalent to minimizing:
\begin{align}
    \mathcal{L}(\mathcal{D}) = \frac{1}{N} \sum_{i=1}^N - \log \pT(\bx^{(i)})
\label{eq:loss1}\end{align}
In case of \emph{continuous} data $\bx$, we minimize the following:
\begin{align}
    \mathcal{L}(\mathcal{D})
    &\simeq \frac{1}{N} \sum_{i=1}^N - \log \pT(\tilde{\bx}^{(i)}) + c
\label{eq:loss2}\end{align}
where $\tilde{\bx}^{(i)} = \bx^{(i)} + u$ with $u \sim \mathcal{U}(0, a)$, and $c = - M \cdot \log a$ where $a$ is determined by the discretization level of the data and $M$ is the dimensionality of $\bx$. Both objectives (eqs.~\eqref{eq:loss1} and \eqref{eq:loss2}) measure the expected compression cost in nats or bits; see ~\citep{dinh2016density}. Optimization is done through stochastic gradient descent using minibatches of data~\citep{kingma2015adam}.

In most flow-based generative models~\citep{dinh2014nice,dinh2016density}, the generative process is defined as:
\begin{align}
\bz &\sim \pT(\bz)\\
\bx &= \gT(\bz)\label{eq:changeofvariables}
\end{align}
where $\bz$ is the latent variable and $\pT(\bz)$ has a (typically simple) tractable density, such as a spherical multivariate Gaussian distribution: $\pT(\bz) = \mathcal{N}(\bz; 0, \mathbf{I})$. The function $\gT(..)$ is invertible, also called \emph{bijective}, such that given a datapoint $\bx$, latent-variable inference is done by $\bz = \fT(\bx) = \gT^{-1}(\bx)$. For brevity, we will omit subscript $\bT$ from $\fT$ and $\gT$.

We focus on functions where $\bbf$ (and, likewise, $\mathbf{g}$) is composed of a sequence of transformations: $\bbf = \bbf_1 \circ \bbf_2 \circ \cdots \circ \bbf_K$, such that the relationship between $\bx$ and $\bz$ can be written as:
\begin{align}
\bx \overset{\bbf_1}{\longleftrightarrow} \bh_1 \overset{\bbf_2}{\longleftrightarrow} \bh_2 \cdots \overset{\bbf_K}{\longleftrightarrow} \bz
\end{align}
Such a sequence of invertible transformations is also called a (normalizing) \emph{flow}~\citep{rezende2015variational}. Under the \emph{change of variables} of eq.~\eqref{eq:changeofvariables}, the probability density function (pdf) of the model given a datapoint can be written as:
\begin{align}
\log \pT(\bx) &= \log \pT(\bz) + \log|\det(d\bz/d\bx)| \\
&= \log \pT(\bz) + \sum_{i=1}^{K} \log|\det(d\bh_i/d\bh_{i-1})|
\end{align}
where we define $\bh_0 \triangleq \bx$ and $\bh_K \triangleq \bz$ for conciseness.
The scalar value $\log|\det(d\bh_i/d\bh_{i-1})|$ is the logarithm of the absolute value of the determinant of the Jacobian matrix $(d\bh_i/d\bh_{i-1})$, also called the \emph{log-determinant}. This value is the change in log-density when going from $\bh_{i-1}$ to $\bh_{i}$ under transformation $\bbf_i$. While it may look intimidating, its value can be surprisingly simple to compute for certain choices of transformations, as previously explored in~\citep{deco1995higher,dinh2014nice,rezende2015variational,kingma2016improving}. The basic idea is to choose transformations whose Jacobian $d\bh_i/d\bh_{i-1}$ is a triangular matrix. For those transformations, the log-determinant is simple:
\SetKwFunction{sum}{sum}
\SetKwFunction{fabs}{abs}
\SetKwFunction{fdiag}{diag}
\begin{align}
 \log|\det(d\bh_i/d\bh_{i-1})| = \sum(\log | \fdiag(d\bh_i/d\bh_{i-1})| )
\end{align}
where $\sum()$ takes the sum over all vector elements, $\log()$ takes the element-wise logarithm, and $\fdiag()$ takes the diagonal of the Jacobian matrix.

\begin{figure*}[t]
    \centering
    \vspace{-0cm}
    \begin{subfigure}[b]{0.45\textwidth}
    \begin{center}
        \centering
        \includegraphics[width=.7\textwidth]{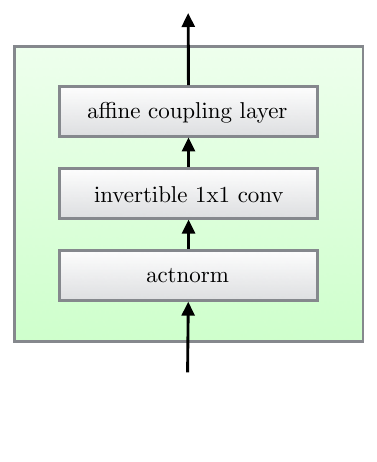}
        \caption{One step of our flow.}
    \end{center}
    \end{subfigure}%
    \vspace{5mm}
    \begin{subfigure}[b]{0.45\textwidth}
        \centering
        \includegraphics[width=.99\textwidth]{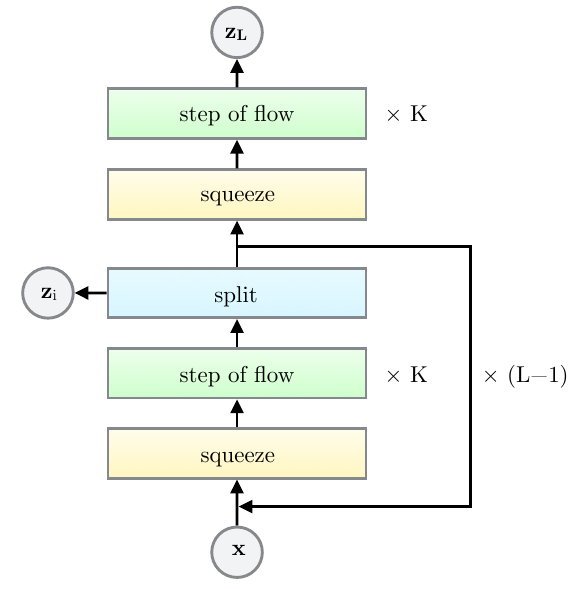}
        \caption{Multi-scale architecture~\citep{dinh2016density}.}
    \end{subfigure}
    \caption{We propose a generative flow where each step (left) consists of an \emph{actnorm} step, followed by an invertible $1\times1$ convolution, followed by an affine transformation~\citep{dinh2014nice}. This flow is combined with a multi-scale architecture (right). See Section~\ref{sec:method} and Table~\ref{functions}.}
    \label{fig:flow}
\end{figure*}

\section{Proposed Generative Flow}\label{sec:method}

We propose a new flow, building on the NICE and RealNVP flows proposed in~\citep{dinh2014nice,dinh2016density}. It consists of a series of steps of flow, combined in a multi-scale architecture; see Figure~\ref{fig:flow}. Each step of flow consists of \emph{actnorm} (Section~\ref{sec:actnorm}) followed by an \emph{invertible $1 \times 1$ convolution} (Section~\ref{sec:invconv}), followed by a coupling layer (Section~\ref{sec:affine}).

This flow is combined with a multi-scale architecture; due to space constraints we refer to ~\citep{dinh2016density} for more details. This architecture has a depth of flow $K$, and number of levels $L$ (Figure~\ref{fig:flow}).

\SetKwFunction{fsplit}{split}
\SetKwFunction{fconcat}{concat}
\SetKwFunction{neuralNet}{NN}
\SetKwFunction{fsqueeze}{squeeze}
\SetKwFunction{funsqueeze}{unsqueeze}
\SetKwFunction{flowStep}{flowStep}
\SetKwFunction{convtwod}{conv2D}
\SetKwFunction{reverseFlowStep}{reverseFlowStep}
\SetKwFunction{return}{return}

\begin{table}
    \smaller
    \caption[The three main components of our proposed flow, their reverses, and their log-determinants]{The three main components of our proposed flow, their reverses, and their log-determinants. Here, $\bx$ signifies the input of the layer, and $\by$ signifies its output. Both $\bx$ and $\by$ are tensors of shape $[h \times w \times c]$ with spatial dimensions $(h, w)$ and channel dimension $c$. With $(i,j)$ we denote spatial indices into tensors $\bx$ and $\by$. The function $\neuralNet()$ is a nonlinear mapping, such as a (shallow) convolutional neural network like in ResNets~\citep{he2016identity} and RealNVP~\citep{dinh2016density}.}
    \label{functions}
  \centering
  \begin{tabularx}{\textwidth}{ X | X | X | l}
    \toprule
    Description & Function & Reverse Function & Log-determinant
    \\\midrule
    \begin{tabular}[t]{@{}l@{}}
    Actnorm.\\
    See Section~\ref{sec:actnorm}.
    \end{tabular}
    &
    $\forall i,j: \by_{i,j} = \bs \odot \bx_{i,j} + \bbb$
    &
    $\forall i,j: \bx_{i,j} = (\by_{i,j} - \bbb) / \bs$
    &
    $h \cdot w \cdot \sum(\log |\bs| )$
    \\\midrule
    \begin{tabular}[t]{@{}l@{}}
    Invertible $1 \times 1$ convolution.\\
    %$\bx : [h \times w \times c]$\\
    $\bW : [c \times c]$.\\
    See Section~\ref{sec:invconv}.
    \end{tabular}
    &
    $\forall i,j: \by_{i,j} = \bW \bx_{i,j}$
    &
    $\forall i,j: \bx_{i,j} = \bW^{-1} \by_{i,j}$
    &
    \begin{tabular}[t]{@{}l@{}}
    $h \cdot w \cdot \log|\det(\bW)|$\\
    or\\
    $h \cdot w \cdot \sum(\log|\bs|)$\\
    (see eq.~\eqref{eq:lu})
    \end{tabular}
    \\\midrule
    \begin{tabular}[t]{@{}l@{}}
    Affine coupling layer.\\
    See Section~\ref{sec:affine} and\\\citep{dinh2014nice}
    \end{tabular}
    &
    \begin{tabular}[t]{@{}l@{}}
    $\bx_a, \bx_b = \fsplit(\bx)$\\
    $(\log \bs, \bt) = \neuralNet(\bx_b)$\\
    $\bs = \exp(\log \bs)$\\
    $\by_a = \bs \odot \bx_a + \bt$\\
    $\by_b = \bx_b$\\
    $\by = \fconcat(\by_a, \by_b)$
    \end{tabular}
    &
    \begin{tabular}[t]{@{}l@{}}
    $\by_a, \by_b = \fsplit(\by)$\\
    $(\log \bs, \bt) = \neuralNet(\by_b)$\\
    $\bs = \exp(\log \bs)$\\
    $\bx_a = (\by_a - \bt) / \bs$\\
    $\bx_b = \by_b$\\
    $\bx = \fconcat(\bx_a, \bx_b)$
    \end{tabular}
    &
    $\sum(\log(|\bs|))$
    %\\\midrule
    %Squeeze: reshapes 3D tensor from $[h,w,c]$ to %$[\frac{1}{2}h,\frac{1}{2}w,4c]$. \citep{dinh2016density}
    %&
    %$\by = \fsqueeze(\bx)$
    %&
    %$\bx = \funsqueeze(\by)$
    %&
    %0
    \\\bottomrule
  \end{tabularx}
\end{table}

% Add to appendix later
% \inputminted[frame=lines,fontsize=\footnotesize]{python}{code.py}

\subsection{Actnorm: scale and bias layer with data dependent initialization}
\label{sec:actnorm}

In \cite{dinh2016density}, the authors propose the use of batch normalization~\citep{ioffe2015batch} to alleviate the problems encountered when training deep models. However, since the variance of activations noise added by batch normalization is inversely proportional to minibatch size per GPU or other processing unit (PU), performance is known to degrade for small per-PU minibatch size. For large images, due to memory constraints, we learn with minibatch size 1 per PU. We propose an \emph{actnorm} layer (for \emph{activation normalizaton}), that performs an affine transformation of the activations using a scale and bias parameter per channel, similar to batch normalization. These parameters are initialized such that the post-actnorm activations per-channel have zero mean and unit variance given an initial minibatch of data. This is a form of \emph{data dependent initialization}~\citep{salimans2016weight}. After initialization, the scale and bias are treated as regular trainable parameters that are independent of the data.

\subsection{Invertible $1 \times 1$ convolution}
\label{sec:invconv}

~\citep{dinh2014nice,dinh2016density} proposed a flow containing the equivalent of a permutation that reverses the ordering of the channels. We propose to replace this fixed permutation with a (learned) invertible $1 \times 1$ convolution, where the weight matrix is initialized as a random rotation matrix. Note that a $1 \times 1$ convolution with equal number of input and output channels is a generalization of a permutation operation.

The log-determinant of an invertible $1 \times 1$ convolution of a $h \times w \times c$ tensor $\bh$ with $c \times c$ weight matrix $\bW$ is straightforward to compute:
\begin{align}
    \log \left| \det \left(\frac{d\,\convtwod(\bh; \bW)}{d\, \bh} \right) \right| = h \cdot w \cdot \log | \det(\bW) |
\end{align}
The cost of computing or differentiating $\det(\bW)$ is $\mathcal{O}(c^3)$, which is often comparable to the cost computing $\convtwod(\bh; \bW)$ which is $\mathcal{O}(h \cdot w \cdot c^2)$. We initialize the weights $\bW$ as a random rotation matrix, having a log-determinant of 0; after one SGD step these values start to diverge from 0.

\paragraph{LU Decomposition.} This cost of computing $\det(\bW)$ can be reduced from $\mathcal{O}(c^3)$ to $\mathcal{O}(c)$ by parameterizing $\bW$ directly in its LU decomposition:
\begin{align}
    \bW = \mathbf{P} \mathbf{L} (\mathbf{U} + \operatorname{diag}(\bs))
\label{eq:lu}\end{align}
where $\mathbf{P}$ is a permutation matrix, $\mathbf{L}$ is a lower triangular matrix with ones on the diagonal, $\mathbf{U}$ is an upper triangular matrix with zeros on the diagonal, and $\bs$ is a vector. The log-determinant is then simply:
\begin{align}
\log | \det(\bW) | = \sum(\log|\bs|)
\end{align}
The difference in computational cost will become significant for large $c$, although for the networks in our experiments we did not measure a large difference in wallclock computation time.

In this parameterization, we initialize the parameters by first sampling a random rotation matrix $\bW$, then computing the corresponding value of $\mathbf{P}$ (which remains fixed) and the corresponding initial values of $\mathbf{L}$ and $\mathbf{U}$ and $\bs$ (which are optimized).

\subsection{Affine Coupling Layers}
\label{sec:affine}

A powerful reversible transformation where the forward function, the reverse function and the log-determinant are computationally efficient, is the \emph{affine coupling} layer introduced in~\citep{dinh2014nice,dinh2016density}. See Table~\ref{functions}. An \emph{additive coupling layer} is a special case with $\bs = 1$ and a log-determinant of 0.

\paragraph{Zero initialization.} We initialize the last convolution of each $\neuralNet()$ with zeros, such that each affine coupling layer initially performs an identity function; we found that this helps training very deep networks.

\paragraph{Split and concatenation.} As in \citep{dinh2014nice}, the $\fsplit()$ function splits $\bh$ the input tensor into two halves along the channel dimension, while the $\fconcat()$ operation performs the corresponding reverse operation: concatenation into a single tensor. In~\citep{dinh2016density}, another type of split was introduced: along the spatial dimensions using a checkerboard pattern. In this work we only perform splits along the channel dimension, simplifying the overall architecture.

\paragraph{Permutation.} Each step of flow above should be preceded by some kind of permutation of the variables that ensures that after sufficient steps of flow, each dimensions can affect every other dimension. The type of permutation specifically done in~\citep{dinh2014nice,dinh2016density} is equivalent to simply \emph{reversing the ordering} of the channels (features) before performing an additive coupling layer. An alternative is to perform a (fixed) random permutation. Our invertible 1x1 convolution is a generalization of such permutations. In experiments we compare these three choices.

\section{Related Work}
\label{sec:relatedwork}

This work builds upon the ideas and flows proposed in ~\citep{dinh2014nice} (NICE) and ~\citep{dinh2016density} (RealNVP); comparisons with this work are made throughout this paper. In~\citep{papamakarios2017masked} (MAF), the authors propose a generative flow based on IAF~\citep{kingma2016improving}; however, since synthesis from MAF is non-parallelizable and therefore inefficient, we omit it from comparisons. Synthesis from autoregressive (AR) models~\citep{hochreiter1997long,graves2013generating,pixelrnn,oord2016conditional,van2016wavenet} is similarly non-parallelizable. Synthesis of high-dimensional data typically takes multiple orders of magnitude longer with AR models; see~\citep{kingma2016improving,oord2017parallel} for evidence. Sampling $256 \times 256$ images with our largest models takes less than one second on current hardware. \footnote{More specifically, generating a $256 \times 256$ image at batch size 1 takes about 130ms on a single 1080 Ti, and about 550ms on a K80}

GANs~\citep{goodfellow2014generative} are arguably best known for their ability to synthesize large and realistic images~\citep{karras2017progressive}, in contrast with likelihood-based methods. Downsides of GANs are their general lack of latent-space encoders, their general lack of full support over the data~\citep{grover2018flow}, their difficulty of optimization, and their difficulty of assessing overfitting and generalization.

%Until a few years ago, the field of deep generative models was dominated by energy-based models such as Restricted Boltzmann Machines (RBM's) \citep{hinton2002training}, or undirected neural networks as as Deep Boltzmann Machines (DBM's) \citep{salakhutdinov2010efficient}. Such models have so far proven difficult to scale due their sensitivity to the \emph{curse of dimensionality}: the difficulty, in high-dimensional spaces, of estimating the partition function and the difficulty of fast-mixing MCMC.

\section{Quantitative Experiments}\label{sec:experiments}

We begin our experiments by comparing how our new flow compares against RealNVP~\citep{dinh2016density}. We then apply our model on other standard datasets and compare log-likelihoods against previous generative models. See the appendix for optimization details. In our experiments, we let each $\neuralNet()$ have three convolutional layers, where the two hidden layers have ReLU activation functions and 512 channels. The first and last convolutions are $3 \times 3$, while the center convolution is $1 \times 1$, since both its input and output have a large number of channels, in contrast with the first and last convolution.

\begin{figure}
	\centering
	\begin{subfigure}{0.5\textwidth}
    	\includegraphics[width=\textwidth]{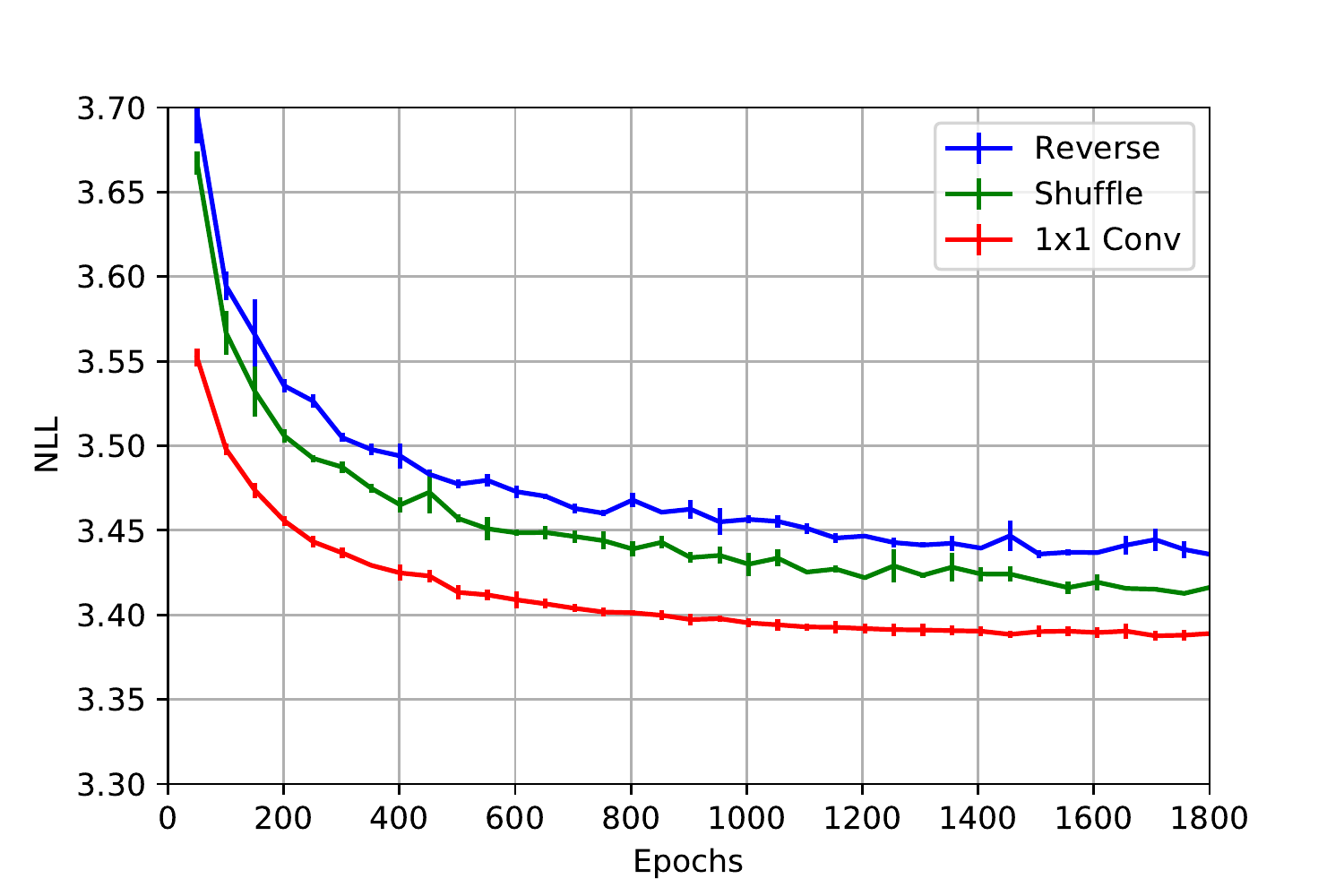}
    	\caption{Additive coupling.}
    \end{subfigure}%
    \begin{subfigure}{0.5\textwidth}
    	\includegraphics[width=\textwidth]{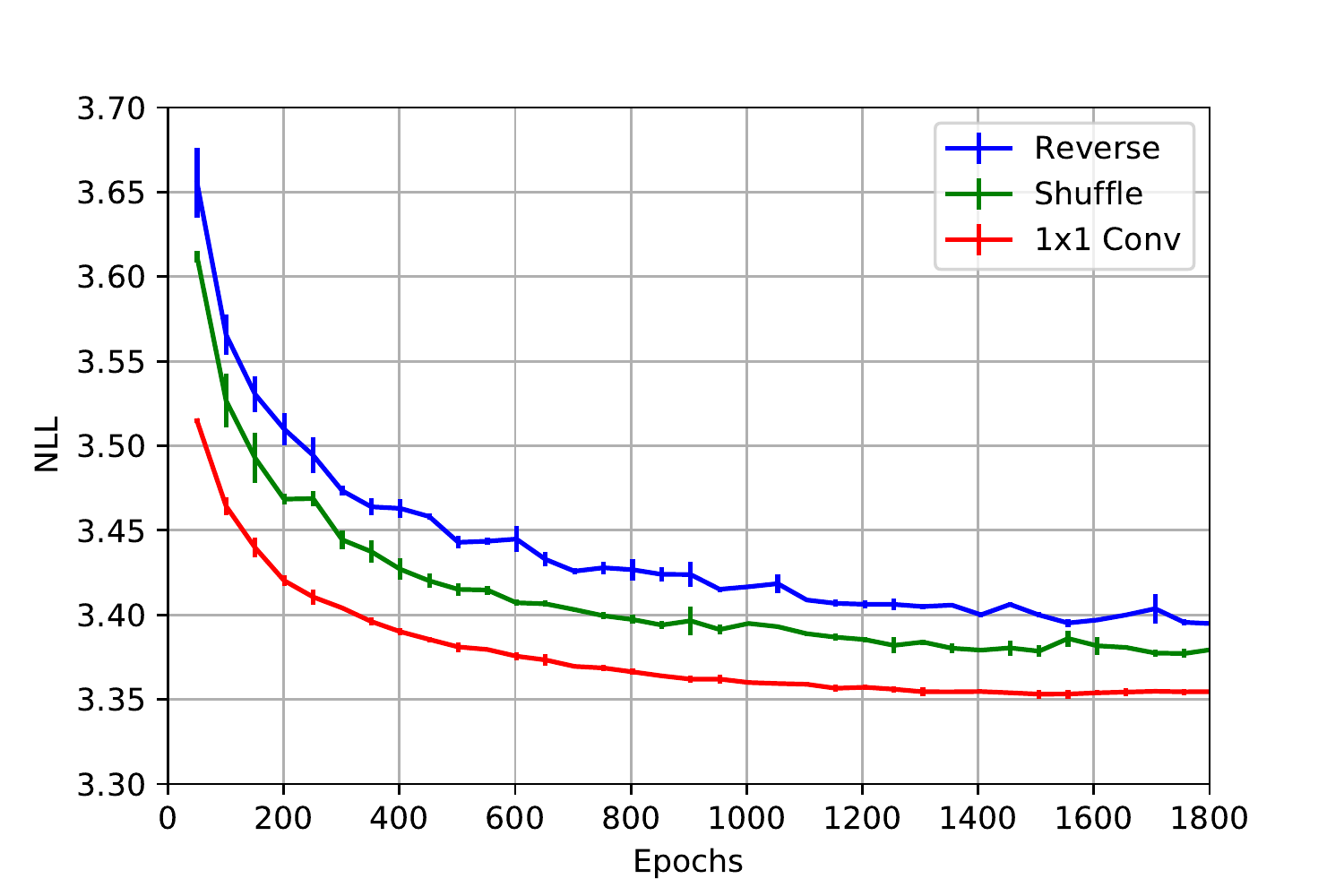}
    	\caption{Affine coupling.}
    \end{subfigure}%
    \caption{Comparison of the three variants - a reversing operation as described in the RealNVP, a fixed random permutation, and our proposed invertible $1 \times 1$ convolution, with additive (left) versus affine (right) coupling layers. We plot the mean and standard deviation across three runs with different random seeds.}
	\label{fig:invconv}
\end{figure}

\paragraph{Gains using invertible $1 \times 1$ Convolution.} We choose the architecture described in Section~\ref{sec:method}, and consider three variations for the permutation of the channel variables - a reversing operation as described in the RealNVP, a fixed random permutation, and our invertible $1 \times 1$ convolution. We compare for models with only additive coupling layers, and models with affine coupling. As described earlier, we initialize all models with a data-dependent initialization which normalizes the activations of each layer. All models were trained with $K=32$ and $L=3$. The model with $1 \times 1$ convolution has a negligible $0.2\%$ larger amount of parameters.

We compare the average negative log-likelihood (bits per dimension) on the CIFAR-10 \citep{krizhevsky2009learning} dataset, keeping all training conditions constant and averaging across three random seeds. The results are in Figure \ref{fig:invconv}. As we see, for both additive and affine couplings, the invertible $1 \times 1$ convolution achieves a lower negative log likelihood and converges faster. The affine coupling models also converge faster than the additive coupling models. We noted that the increase in wallclock time for the invertible $1 \times 1$ convolution model was only $\approx 7\%$, thus the operation is computationally efficient as well.

\paragraph{Comparison with RealNVP on standard benchmarks.}
\label{sec:exp_ll}
Besides the permutation operation, the RealNVP architecture has other differences such as the spatial coupling layers. In order to verify that our proposed architecture is overall competitive with the RealNVP architecture, we compare our models on various natural images datasets. In particular, we compare on CIFAR-10, ImageNet \citep{russakovsky2015imagenet} and LSUN \citep{yu15lsun} datasets. We follow the same preprocessing as in \citep{dinh2016density}. For Imagenet, we use the $32 \times 32$ and $64 \times 64$ downsampled version of ImageNet ~\citep{oord2016pixel}, and for LSUN we downsample to $96\times96$ and take random crops of $64\times64$. We also include the bits/dimension for our model trained on $256 \times 256$ CelebA HQ used in our qualitative experiments.\footnote{Since the original CelebA HQ dataset didn't have a validation set, we separated it into a training set of 27000 images and a validation set of 3000 images} As we see in Table \ref{tab:models:results}, our model achieves a significant improvement on all the datasets.

\begin{table}[]
    \centering
    \caption{Best results in bits per dimension of our model compared to RealNVP.}
    \label{tab:models:results}
    \resizebox{\textwidth}{!}{%
    \begin{tabular}{l | l | l | l | l | l | l}
    \toprule
    Model     & CIFAR-10 & ImageNet 32x32 & ImageNet 64x64 & LSUN (bedroom) & LSUN (tower) & LSUN (church outdoor)
    \\\midrule
    RealNVP & $3.49$ & $4.28$ & $3.98$ & 2.72 & 2.81 & 3.08
    \\\midrule
    Glow    & $\mathbf{3.35}$ & $\mathbf{4.09}$ & $\mathbf{3.81}$ &  $\mathbf{2.38}$ &  $\mathbf{2.46}$ & $\mathbf{2.67}$
    \\\bottomrule
    \end{tabular}}
\end{table}

\begin{figure}[t]
	\centering
	\includegraphics[width=0.8\textwidth]{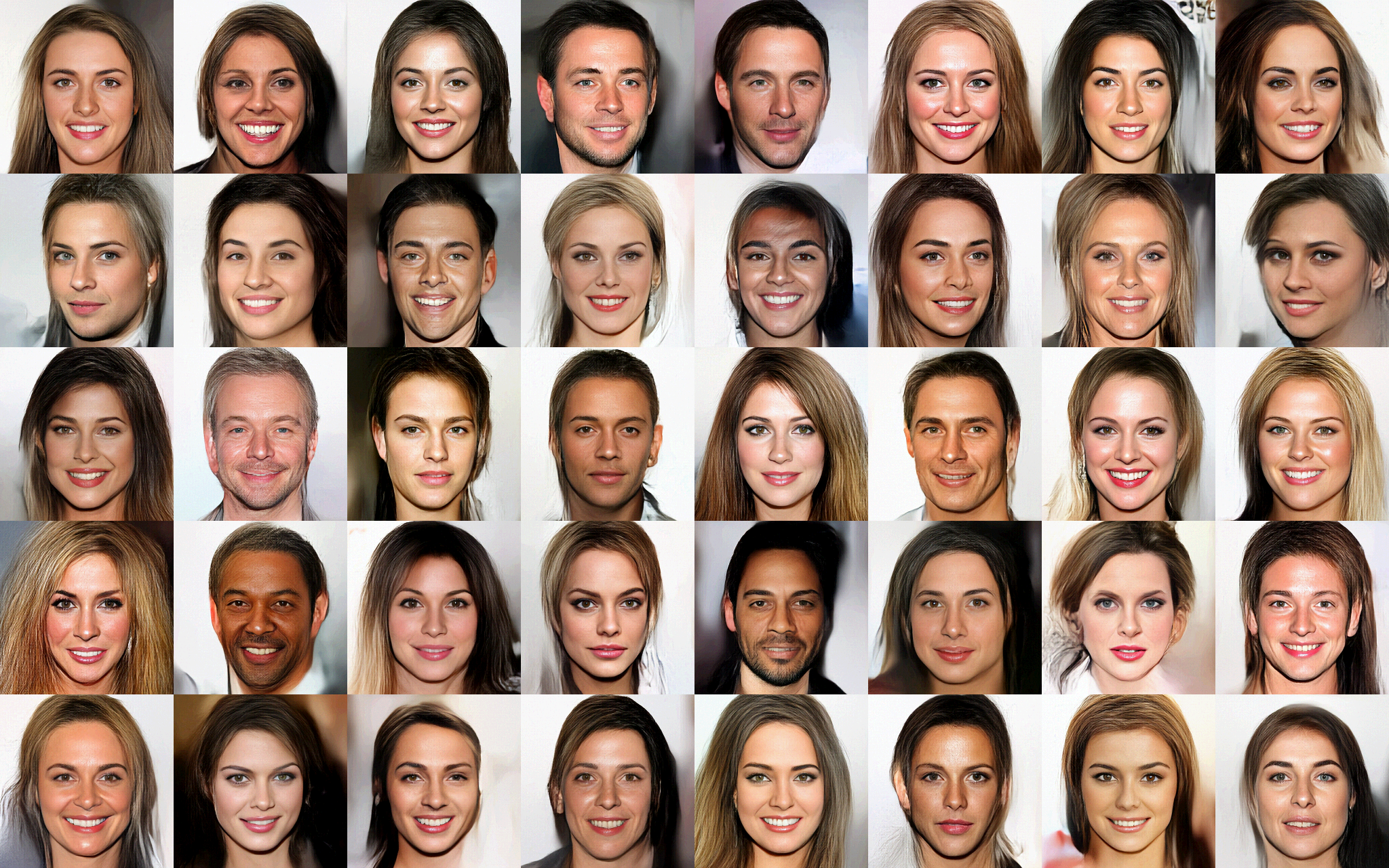}
	\caption{Random samples from the model, with temperature $0.7$}
	\label{fig:samples}
\end{figure}
\begin{figure}[t]
	\centering
	\includegraphics[width=0.8\textwidth]{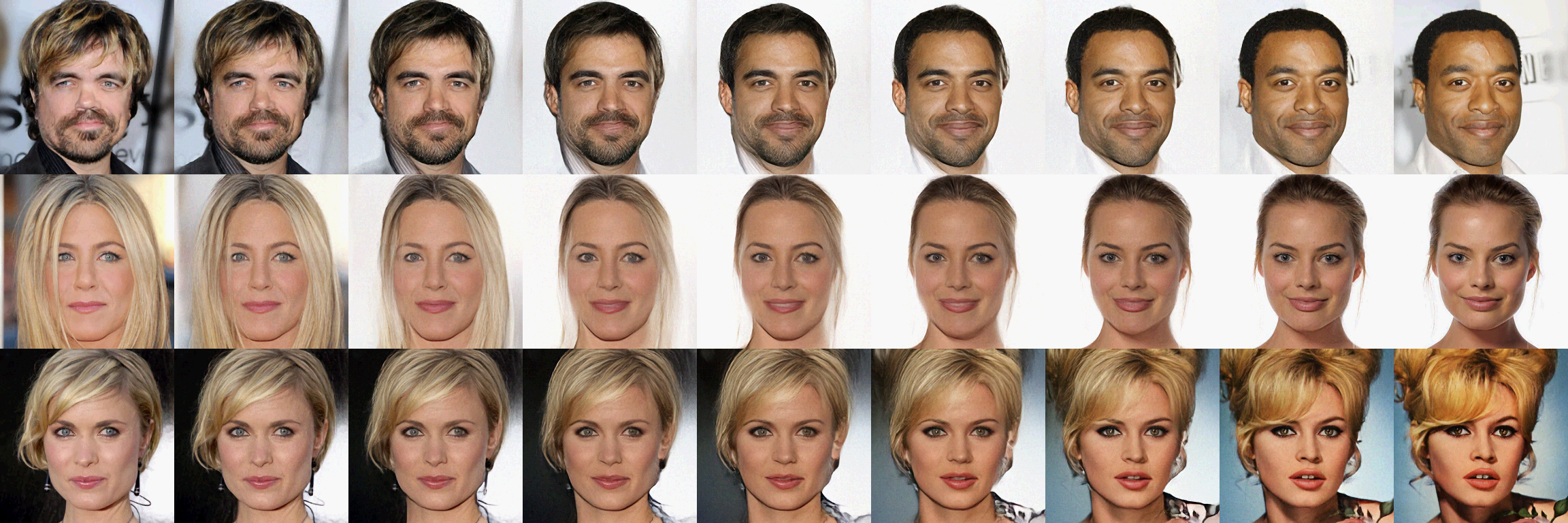}
	\caption{Linear interpolation in latent space between real images}
	\label{fig:interpolation}
\end{figure}

\section{Qualitative Experiments}\label{sec:qualexperiments}
We now study the qualitative aspects of the model on high-resolution datasets. We choose the CelebA-HQ dataset \citep{karras2017progressive}, which consists of $30000$ high resolution images from the CelebA dataset, and train the same architecture as above but now for images at a resolution of $256^2$, $K=32$ and $L=6$. To improve visual quality at the cost of slight decrease in color fidelity, we train our models on $5$-bit images. We aim to study if our model can scale to high resolutions, produce realistic samples, and produce a meaningful latent space. Due to device memory constraints, at these resolutions we work with minibatch size 1 per PU, and use gradient checkpointing \citep{gradcheckpointing}. In the future, we could use a constant amount of memory independent of depth by utilizing the reversibility of the model \citep{gomez2017reversible}.

Consistent with earlier work on likelihood-based generative models~\citep{parmar2018image}, we found that sampling from a reduced-temperature model often results in higher-quality samples. When sampling with temperature $T$, we sample from the distribution $p_{\boldsymbol{\theta},T}(\bx) \propto (\pT(\bx))^{T^2}$. In case of additive coupling layers, this can be achieved simply by multiplying the standard deviation of $\pT(\bz)$ by a factor of $T$.

\paragraph{Synthesis and Interpolation.}
Figure \ref{fig:samples} shows the random samples obtained from our model. The images are extremely high quality for a non-autoregressive likelihood based model. To see how well we can interpolate, we take a pair of real images, encode them with the encoder, and linearly interpolate between the latents to obtain samples. The results in Figure \ref{fig:interpolation} show that the image manifold of the generator distribution is extremely smooth and almost all intermediate samples look like realistic faces.

\paragraph{Semantic Manipulation.}
\begin{figure}[t]
    \centering
    \begin{subfigure}{0.45\textwidth}
        \centering
        \includegraphics[width=\textwidth]{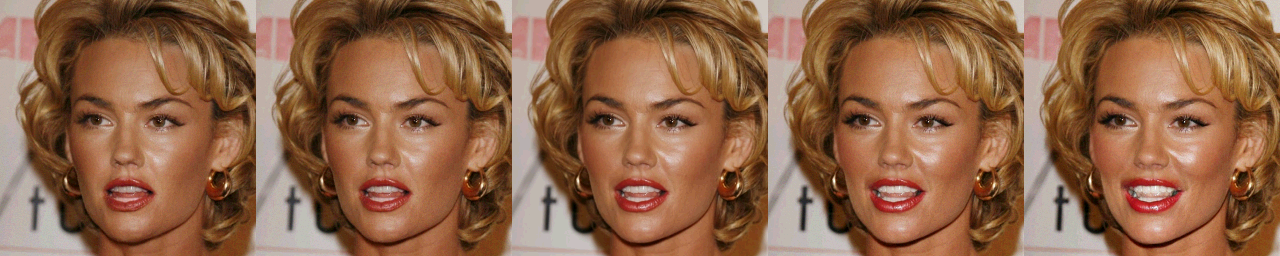}
        \caption{Smiling}
    \end{subfigure}
    \begin{subfigure}{0.45\textwidth}
        \centering
        \includegraphics[width=\textwidth]{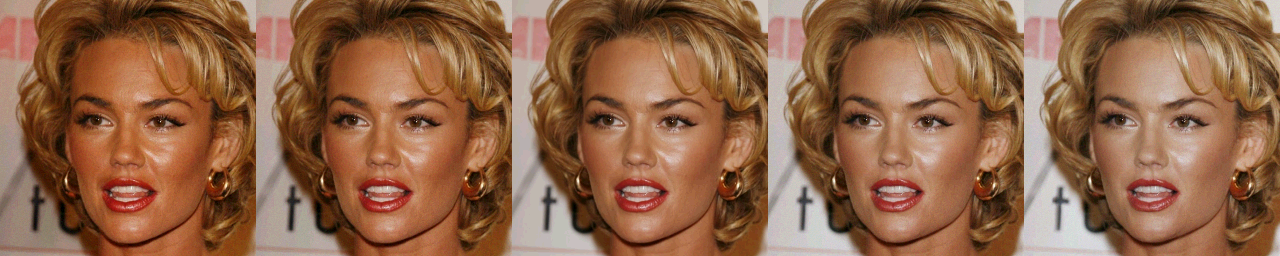}
        \caption{Pale Skin}
    \end{subfigure}
    \begin{subfigure}{0.45\textwidth}
        \centering
        \includegraphics[width=\textwidth]{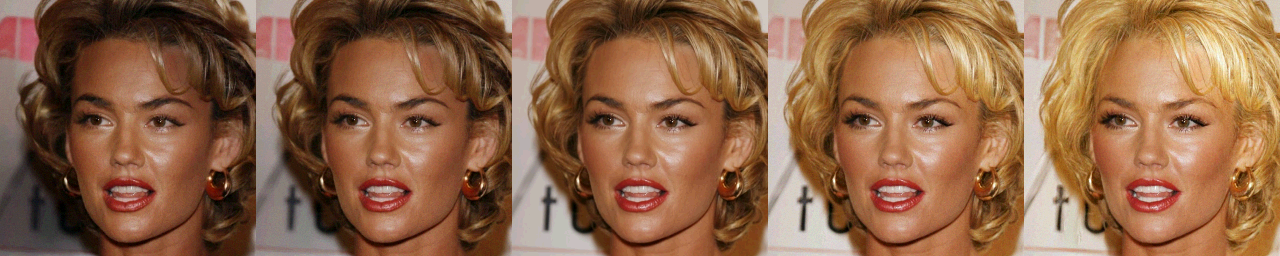}
        \caption{Blond Hair}
    \end{subfigure}
    \begin{subfigure}{0.45\textwidth}
        \centering
        \includegraphics[width=\textwidth]{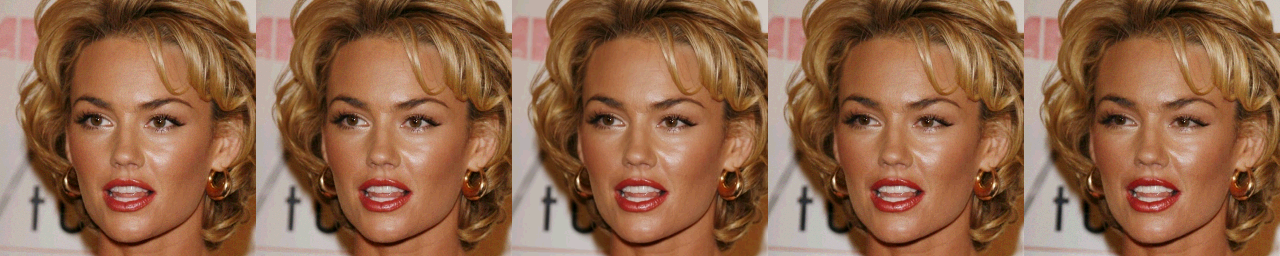}
        \caption{Narrow Eyes}
    \end{subfigure}
    \begin{subfigure}{0.45\textwidth}
        \centering
        \includegraphics[width=\textwidth]{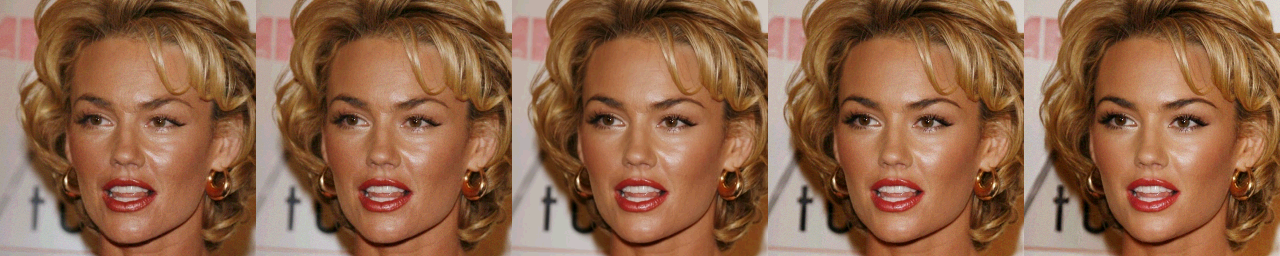}
        \caption{Young}
    \end{subfigure}
    \begin{subfigure}{0.45\textwidth}
        \centering
        \includegraphics[width=\textwidth]{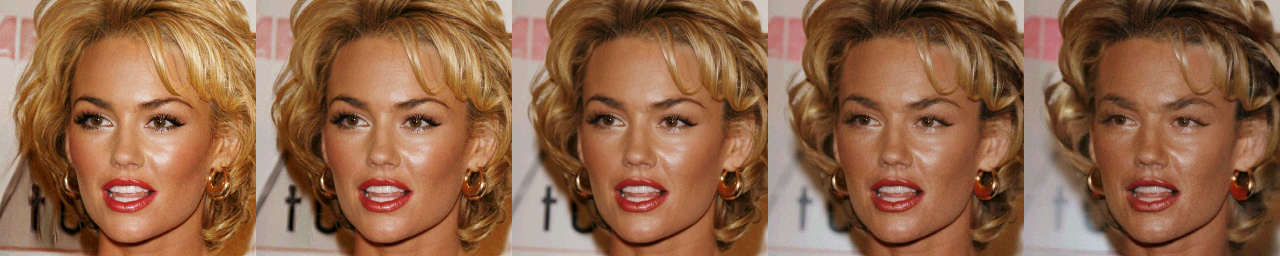}
        \caption{Male}
    \end{subfigure}
    \caption{Manipulation of attributes of a face. Each row is made by interpolating the latent code of an image along a vector corresponding to the attribute, with the middle image being the original image}
    \label{fig:manipulation}
\end{figure}
We now consider modifying attributes of an image. To do so, we use the labels in the CelebA dataset. Each image has a binary label corresponding to presence or absence of attributes like smiling, blond hair, young, etc. This gives us $30000$ binary labels for each attribute. We then calculate the average latent vector $\bz_{pos}$ for images with the attribute and $\bz_{neg}$ for images without, and then use the difference $(\bz_{pos} - \bz_{neg})$ as a direction for manipulating. Note that this is a relatively small amount of supervision, and is done after the model is trained (no labels were used while training), making it extremely easy to do for a variety of different target attributes. The results are shown in Figure \ref{fig:manipulation}.

\begin{figure}
	\centering
	\begin{subfigure}{0.32\textwidth}
    	\includegraphics[width=\textwidth]{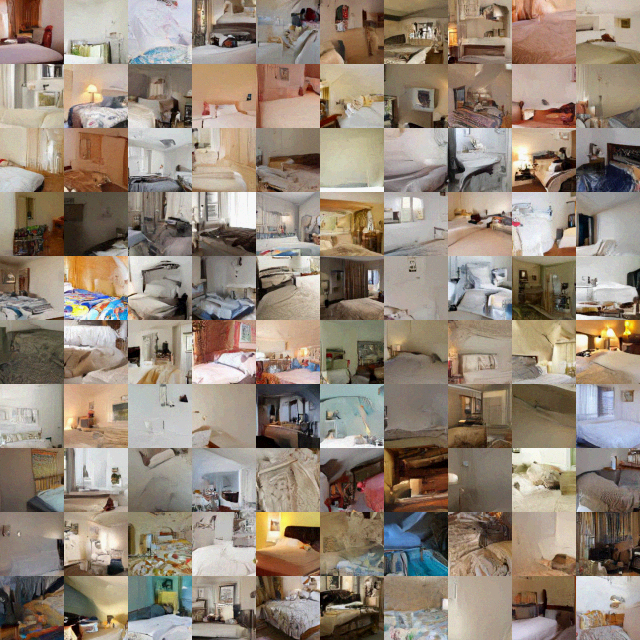}
    \end{subfigure}\hspace{0.01\textwidth}%
	\begin{subfigure}{0.32\textwidth}
    	\includegraphics[width=\textwidth]{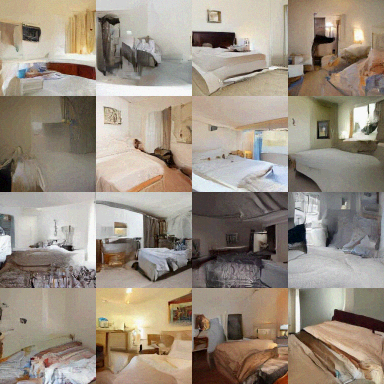}
    \end{subfigure}\hspace{0.01\textwidth}%
    \begin{subfigure}{0.32\textwidth}
    	\includegraphics[width=\textwidth]{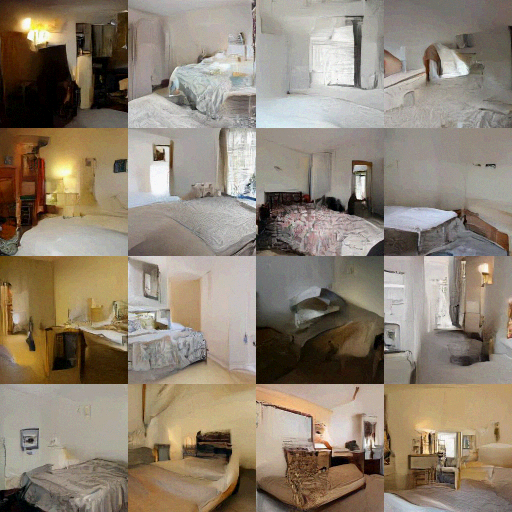}
    \end{subfigure}%
    \caption[Samples from model trained on 5-bit LSUN bedrooms, at temperature 0.875. Resolutions 64, 96 and 128 respectively]{Samples from model trained on 5-bit LSUN bedrooms, at temperature 0.875. Resolutions 64, 96 and 128 respectively \footnotemark}
\end{figure}

\begin{figure}
	\centering
	\includegraphics[width=0.6\textwidth]{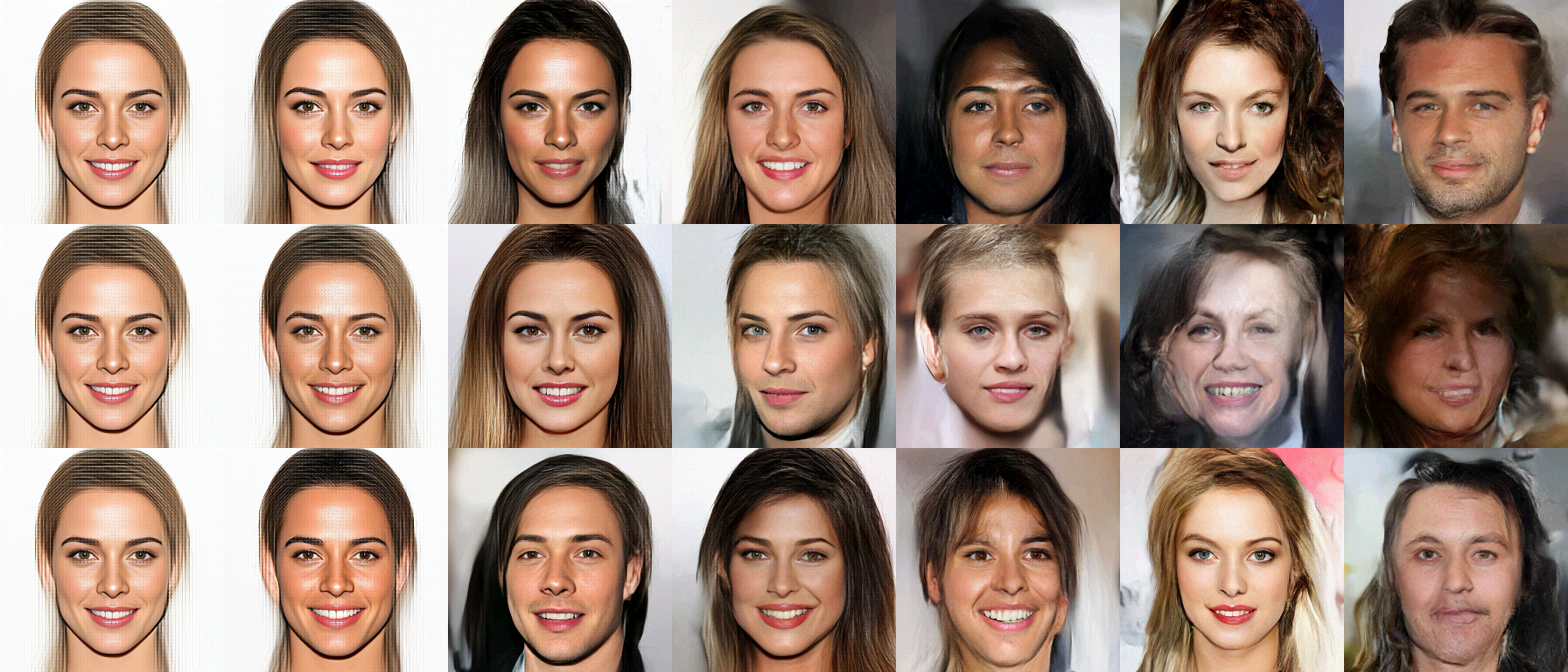}
	\caption{Effect of change of temperature. From left to right, samples obtained at temperatures $0, 0.25, 0.6, 0.7, 0.8, 0.9, 1.0$}
	\label{fig:temperature}
\end{figure}
\begin{figure}
	\centering
	\includegraphics[width=0.6\textwidth]{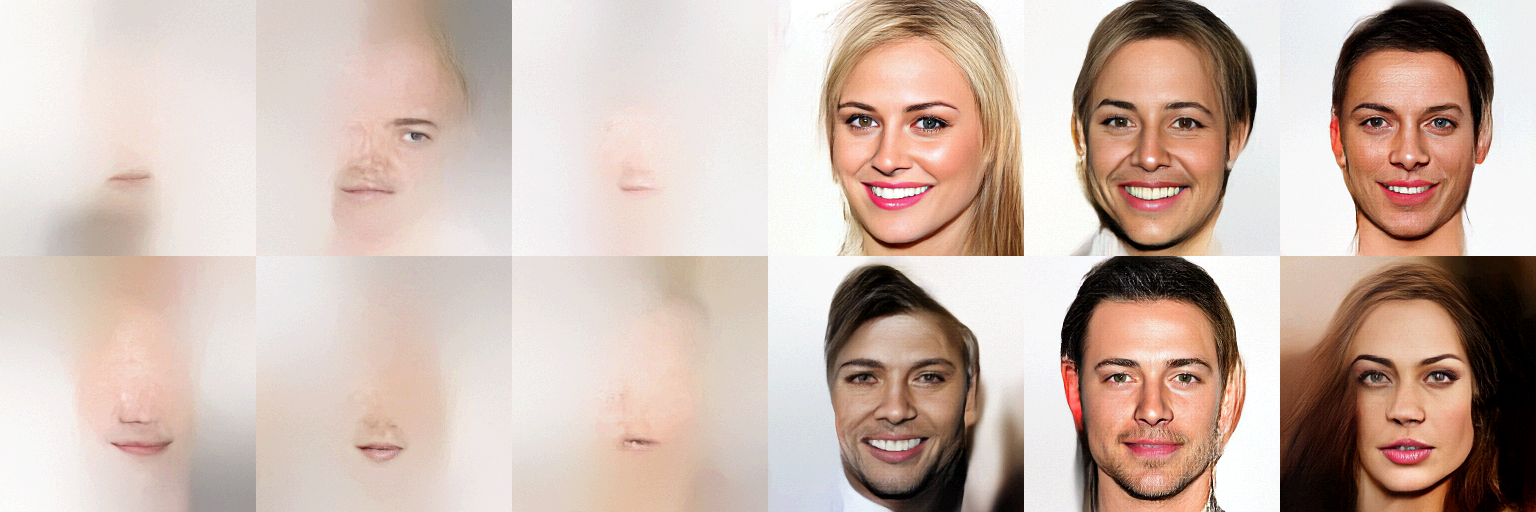}
	\caption{Samples from shallow model on left vs deep model on right. Shallow model has $L = 4$ levels, while deep model has $L = 6$ levels}
	\label{fig:depth}
\end{figure}

\paragraph{Effect of temperature and model depth.} Figure \ref{fig:temperature} shows how the sample quality and diversity varies with temperature. The highest temperatures have noisy images, possibly due to overestimating the entropy of the data distribution, and thus we choose a temperature of $0.7$ as a sweet spot for diversity and quality of samples. Figure \ref{fig:depth} shows how model depth affects the ability of the model to learn long-range dependencies.

\begin{comment}
 - Show random samples from face model
 - Show interpolations between faces
 - Show samples of not-deep-enough model with the deep model
 - Show manipulations of faces
\end{comment}

\footnotetext{For $128 \times 128$ and $96 \times 96$ versions, we centre cropped the original image, and downsampled. For $64 \times 64$ version, we took random crops from the $96 \times 96$ downsampled image as done in \cite{dinh2016density}}
\section{Conclusion}

We propose a new type of flow, coined~\emph{Glow}, and demonstrate improved quantitative performance in terms of log-likelihood on standard image modeling benchmarks. In addition, we demonstrate that when trained on high-resolution faces, our model is able to synthesize realistic images. Our model is, to the best of our knowledge, the first likelihood-based model in the literature that can efficiently synthesize high-resolution natural images.

\smaller
\bibliographystyle{apalike}
\bibliography{bib}

% Hi! We'll have to submit supplementary material seaprately, so I'll move this to a separate tex file / pdf? supp.tex

% OK! Done
% \newpage

\appendix
\normalsize

\section{Additional quantitative results}
See Table~\ref{tab:models:moreresults}.
\begin{table}[h]
    \centering
    \caption{Quantiative results in bits per dimension on the test set.}
    \label{tab:models:moreresults}
    \begin{tabular}{l | l}
    \toprule
    Dataset & Glow
    \\\midrule
    CIFAR-10, 32$\times$32, 5-bit & 1.67
    \\\midrule
    ImageNet, 32$\times$32, 5-bit & 1.99
    \\\midrule
    ImageNet, 64$\times$64, 5-bit & 1.76
    \\\midrule
    CelebA HQ, 256$\times$256, 5-bit & 1.03
    \\\bottomrule
    \end{tabular}
\end{table}

\section{Simple python implementation of the invertible $1 \times 1$ convolution}
% \inputminted[frame=lines,fontsize=\footnotesize]{python}{code.py}
\begin{lstlisting}[language=Python]
# Invertible 1x1 conv
def invertible_1x1_conv(z, logdet, forward=True):
    # Shape
    h,w,c = z.shape[1:]

    # Sample a random orthogonal matrix to initialise weights
    w_init = np.linalg.qr(np.random.randn(c,c))[0]
    w = tf.get_variable("W", initializer=w_init)

    # Compute log determinant
    dlogdet = h * w * tf.log(abs(tf.matrix_determinant(w)))

    if forward:
        # Forward computation
        _w = tf.reshape(w, [1,1,c,c])
        z = tf.nn.conv2d(z, _w, [1,1,1,1], 'SAME')
        logdet += dlogdet

        return z, logdet
    else:
        # Reverse computation
        _w = tf.matrix_inverse(w)
        _w = tf.reshape(_w, [1,1,c,c])
        z = tf.nn.conv2d(z, _w, [1,1,1,1], 'SAME')
        logdet -= dlogdet

        return z, logdet
\end{lstlisting}
\section{Optimization details}\label{sec:opt}

We use the Adam optimizer~\citep{kingma2015adam} with $\alpha=0.001$ and default $\beta_1$ and $\beta_2$. In out quantitative experiments (Section \ref{sec:experiments}, Table \ref{tab:models:results}) we used the following hyperparameters (Table \ref{tab:models:hyper}). \\
\begin{table}[h]
    \centering
    \caption{Hyperparameters for results in Section \ref{sec:experiments}, Table \ref{tab:models:results}}
    \label{tab:models:hyper}
    \begin{tabular}{l | l | l | l | l}
    \toprule
    Dataset & Minibatch Size & Levels (L) & Depth per level (K) & Coupling
    \\\midrule
    CIFAR-10 & 512 & 3 & 32 & Affine
    \\\midrule
    ImageNet, 32$\times$32 & 512 & 3 & 48 & Affine
    \\\midrule
    ImageNet, 64$\times$64 & 128 & 4 & 48 & Affine
    \\\midrule
    LSUN, 64$\times$64 & 128 & 4 & 48 & Affine
    \\\bottomrule
    \end{tabular}
\end{table}

In our qualitative experiments (Section \ref{sec:qualexperiments}), we used the following hyperparameters (Table \ref{tab:models:hyper_qual})
\begin{table}[h]
    \centering
    \caption{Hyperparameters for results in Section \ref{sec:qualexperiments}}
    \label{tab:models:hyper_qual}
    \begin{tabular}{l | l | l | l | l}
    \toprule
    Dataset & Minibatch Size & Levels (L) & Depth per level (K) & Coupling
    \\\midrule
    LSUN, 64$\times$64, 5-bit & 128 & 4 & 48 & Additive
    \\\midrule
    LSUN, 96$\times$96, 5-bit & 320 & 5 & 64 & Additive
    \\\midrule
    LSUN, 128$\times$128, 5-bit & 160 & 5 & 64 & Additive
    \\\midrule
    CelebA HQ, 256$\times$256, 5-bit & 40 & 6 & 32 & Additive
    \\\bottomrule
    \end{tabular}
\end{table}
\section{Extra samples from qualitative experiments}
\begin{figure}[h]
	\centering
	\begin{subfigure}{0.69\textwidth}
    	\includegraphics[width=\textwidth]{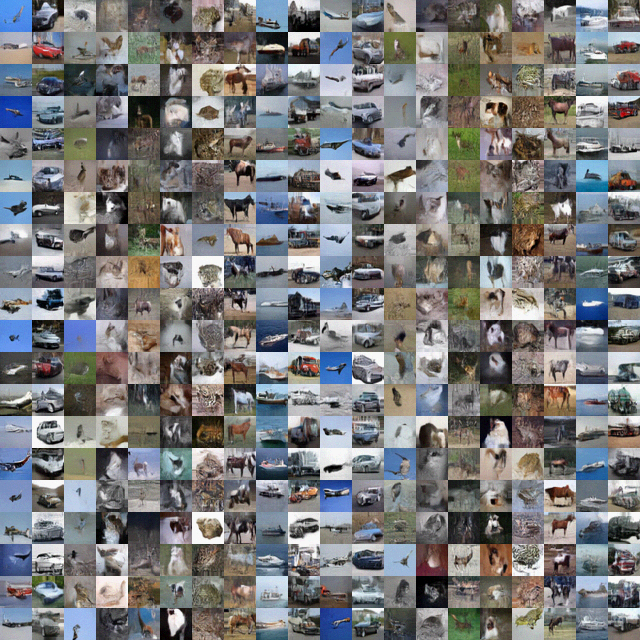}
	\caption{Class conditional CIFAR-10 samples}
	\end{subfigure}\hspace{0.01\textwidth}%
	\begin{subfigure}{0.69\textwidth}
    	\includegraphics[width=\textwidth]{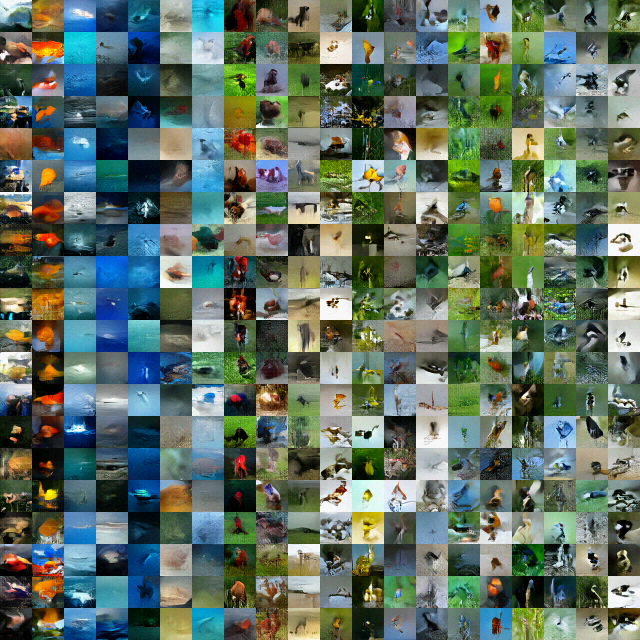}
	\caption{Class conditional $32 \times 32$ ImageNet samples}
	\end{subfigure}
\caption{Class conditional samples on 5-bit CIFAR-10 and $32 \times 32$ ImageNet respectively. Temperature 0.75}
\label{fig:cif_img_qual}
\end{figure}

For the class conditional CIFAR-10 and 32$\times$32 ImageNet samples, we used the same hyperparameters as the quantitative experiments, but with a class dependent prior at the top-most level. We also added a classification loss to predict the class label from the second last layer of the encoder, with a weight of $\lambda = 0.01$. The results are in Figure \ref{fig:cif_img_qual}.

\section{Extra samples from the quantitative experiments}
\begin{figure}[h]
	\centering
	\begin{subfigure}{0.49\textwidth}\includegraphics[width=\textwidth]{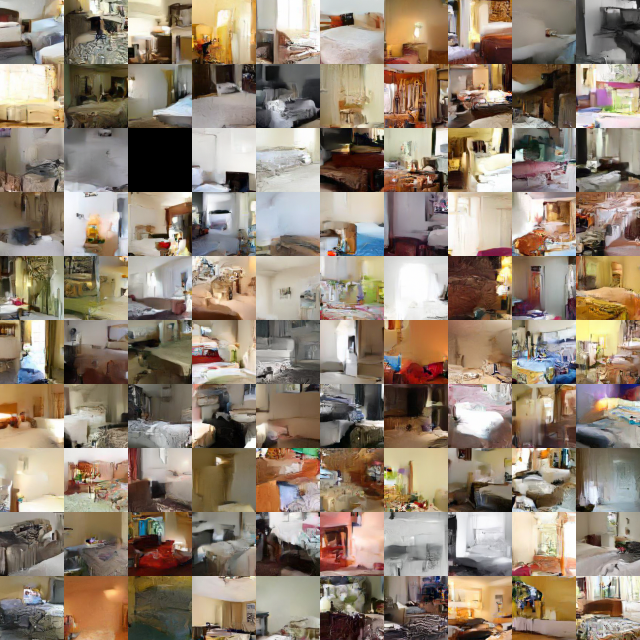}
	\end{subfigure}\hspace{0.01\textwidth}%
	\begin{subfigure}{0.49\textwidth}\includegraphics[width=\textwidth]{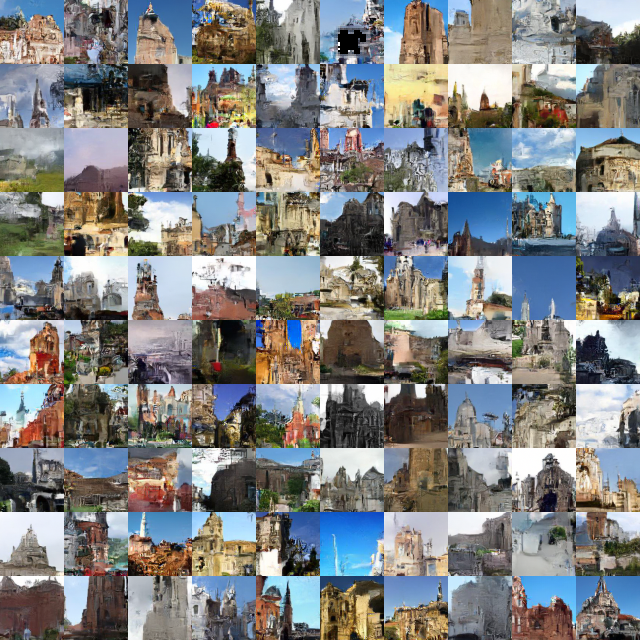}
	\end{subfigure}\hspace{0.01\textwidth}%
	\begin{subfigure}{0.49\textwidth}\includegraphics[width=\textwidth]{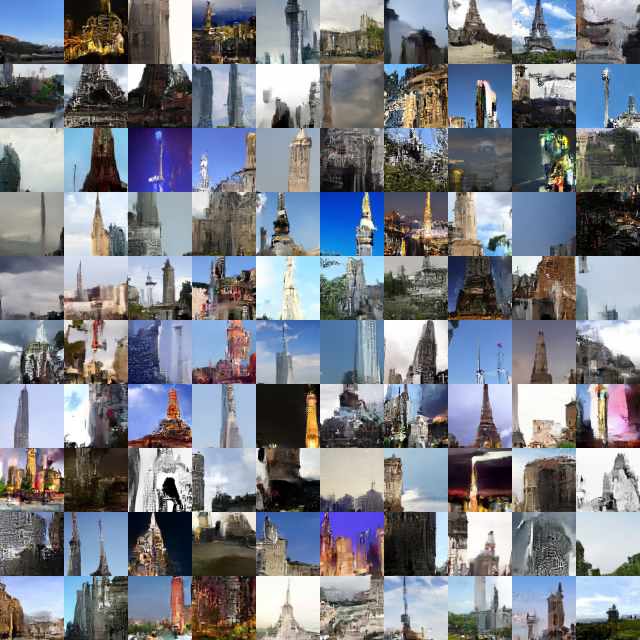}
	\end{subfigure}%
\caption{Samples from 8-bit, $64 \times 64$ LSUN bedrooms, church and towers respectively. Temperature 1.0}
\label{fig:lsun_quant}
\end{figure}

\begin{figure}
    \centering
    \includegraphics[width=0.69\textwidth]{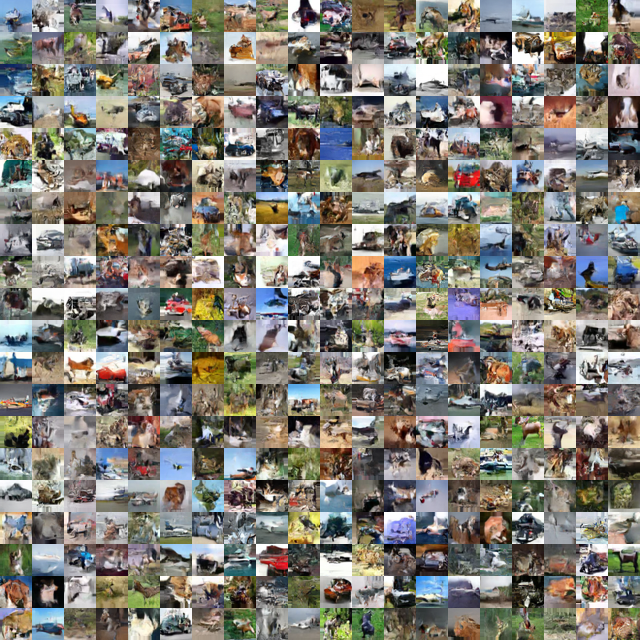}
    \caption{Samples from an unconditional model with affine coupling layers trained on the CIFAR-10 dataset with temperature 1.0.}
    \label{fig:my_label}
\end{figure}

For direct comparison with other work, datasets are preprocessed exactly as in \cite{dinh2016density}. Results are in Figure \ref{fig:lsun_quant} and Figure \ref{fig:my_label}.

\end{document}